\documentclass{sig-alternate-05-2015}
\usepackage[sc]{mathpazo}
\usepackage[T1]{fontenc}

\usepackage[keeplastbox]{flushend}

\usepackage{hyperref}
\usepackage[usenames,dvipsnames]{color}
\usepackage{booktabs}
\usepackage{graphicx}
\usepackage{subfigure}
\usepackage[normalem]{ulem}
\usepackage[labelfont=bf,font=small]{caption}

\newcommand{\sref}[1]{Sec. \ref{#1}}
\newcommand{\figref}[1]{Fig.\ref{#1}}

\newcommand{\adnote}[1]%
 {\textcolor{blue}{\textbf{AD: #1}}}
 
 \newcommand{\prg}[1]{\noindent\textbf{#1. }} 
\newcommand{\be}[1]{\textbf{\emph{#1}}}

\begin{document}

\setcopyright{none}

\pagestyle{plain} 

%

\title{Do You Want Your Autonomous Car To Drive Like You?}
%
%
%
%
%

\numberofauthors{5} 
%

 \author{
 \alignauthor
 Chandrayee Basu\\
       \affaddr{UC Merced}\\
       \email{cbasu2@ucmerced.edu}
 \alignauthor
 Qian Yang\\
       \affaddr{Carnegie Mellon University}\\
       \email{qyang1@cs.cmu.edu}
 \alignauthor
 David Hungerman\\
       \affaddr{UC Merced}\\
       \email{dmhungerman@ucmerced.edu}  
\and
 \alignauthor
  Mukesh Singhal\\
       \affaddr{UC Merced}\\
       \email{msinghal@ucmerced.edu}
 \alignauthor
 Anca D. Dragan\\
       \affaddr{UC Berkeley}\\
       \email{anca@berkeley.edu}
      }



\maketitle
\begin{abstract}

With progress in enabling autonomous cars to drive safely on the road, it is time to start asking \emph{how} they should be driving. A common answer is that they should be adopting their users' driving style. This makes the assumption that users want their autonomous cars to drive like they drive -- aggressive drivers want aggressive cars, defensive drivers want defensive cars. In this paper, we put that assumption to the test. We find that users tend to prefer a significantly more defensive driving style than their own. Interestingly, they prefer the style they \emph{think} is their own, even though their actual driving style tends to be more aggressive. We also find that preferences do depend on the specific driving scenario, opening the door for new ways of learning driving style preference.   

\end{abstract}

%
%


%
%

%
%


\keywords{autonomous cars; driving preferences; driving style}

\section{Introduction}
In the age of autonomous driving, researchers and companies are getting ever-so-close to enabling cars to generate driving behavior that includes reaching the destination while satisfying safety constraints, like not colliding with other cars or pedestrians. 

Once autonomous cars attain that level of capability, initially, they might be able to generate, for each driving situation, only one solution trajectory (or behavior) that satisfies these safety and feasibility constraints. But really, many solutions exist -- there are many ways to drive. This depend on the individual trade-offs that each driver makes. We have an existence proof for that. Some of us are more \emph{aggressive} drivers, valuing efficiency and being comfortable getting close to other cars on the road. Others are more \emph{defensive}, a bit more conservative when it comes to safety, leaving a large distance to the next car for example, or quickly braking when someone attempts to merge in front. 

Soon after we are able to generate \emph{one} feasible behavior, we will be asking ourselves \emph{which} behavior we should try to generate: what driving \emph{style} should an autonomous car have? There is a natural answer to this question: cars should do what users want them to \cite{kuderer2015learning, Here360, karjanto2015comfortstyle}. If the user drives aggressively, so should the car. The car should borrow the user's driving style (though not the imperfections). This is very apparent from the expression ``back seat driving'', which suggests that people want the driver to do what they would do.

\begin{figure}[t!]
    \centering
    \includegraphics[width=\columnwidth]{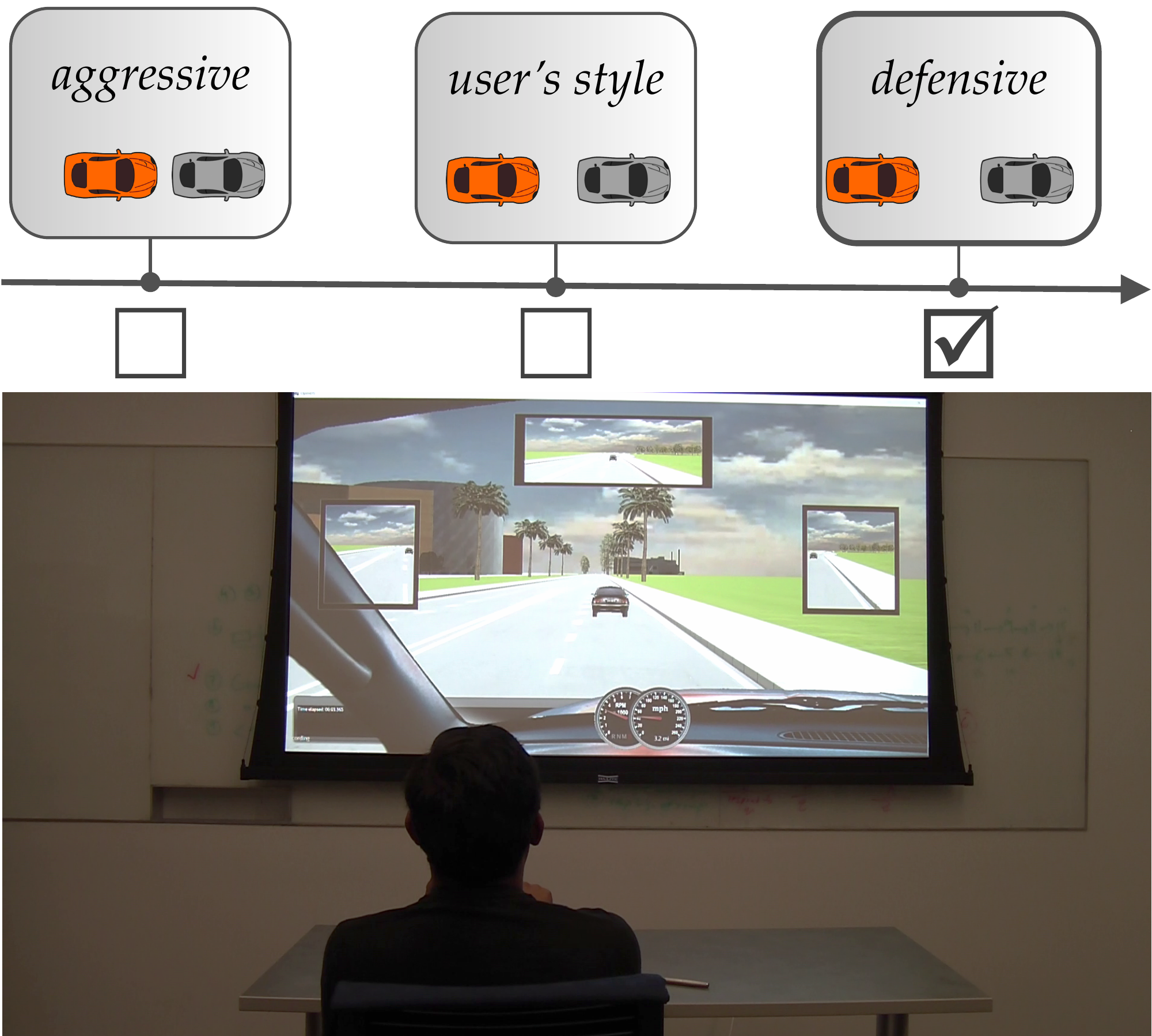}
    \caption{We first get data from user driving in different scenarios, and in a second session ask them to compare their own style (without knowing it is theirs), a more defensive style, and a more aggressive style. Participants tended to prefer a more defensive style than their own, but mistakenly thought they were actually picking their own. }
    \label{fig:set up}
\end{figure}

Prior work has focused on identifying the user's driving style, via Inverse Reinforcement Learning \cite{kuderer2015learning,sadighinformation,lam2015efficient}.  
In all of them, the underlying assumption is that we want cars to match our driving style: that we want them to drive like us.

In this paper, we challenge this assumption, and hypothesize that users want a driving style that is different from their own. We design and conduct a user study to start analyzing the potential differences between how users drive and how they want to be driven. Our study, conducted in a driving simulator, has two parts: first, the users come in and demonstrate their driving in different environments; second, at a later date, the same users come in and test four driving styles: their own (though they do not know it is their own), an aggressive style, a defensive style, and another user's style. We measure their preference for these styles, as well as the perceived similarity to their own style.

Our results suggest that there is truth to both sides:
\begin{quote}
Users do not actually want the car to drive like they drive. Instead, they want the car to drive like they \emph{think} they drive. 
\end{quote}

We found a significant difference in user's own style and preferred style, with users typically preferring more defensive driving when they are passengers. However, we also found a strong correlation between the style that users preferred, and the style that users perceived as closest to their own. There was little correlation, however, between what they thought was their own style and what \emph{actually} was their own style. 

Overall, our work does not contradict the need for customization, but suggests that it might not be sufficient to learn how the user drives. Instead, we need to learn how the user actually wants to be driven. This raises challenges for learning, because we can no longer rely on demonstrations -- users can easily demonstrate how they drive, but they might not be able to demonstrate the driving style they want. Instead, we need to rely on different kinds of input and guidance from users in the learning process. 

Furthermore, there is a tension between what users think they want (their style) and what they actually want (a more defensive style). On the brighter side, our results suggest that the learned output should be easily accepted by users: when the car drives in the preferred style, chances are users will perceive it as their own  style anyway!

We define \textit{driving style}, informed by prior work, in Related Work, followed by our statement of hypothesis, description of the manipulated variables, the simulation environment, the user studies and the confounds in the Methods section. Here, we also present a quantitative measure of driving style in terms of driving features derived from prior research. The rest of paper is organized into Results and Discussion.

\section{Related Work}
The typical behavioral patterns of a driver are usually referred to by the term \emph{driving style}. This includes the choice of driving speed, headway, overtaking of other vehicles, or the tendency to commit traffic violations \cite{van2015measuring}.

Defensiveness-aggressiveness is the most commonly used metric for defining driving style. Prior work refers to drivers as aggressive/assertive versus defensive \cite{karjanto2015comfortstyle}; or mild versus moderate versus aggressive \cite{xu2015establishing}. In the Multidimensional Driving Style Inventory (MDSI), Taubman-Ben-Ari \textit{et al.} identified four broad driving styles: (1) reckless and careless driving, characterized by, for example, higher speed; (2) anxious driving; (3) angry and hostile driving, characterized by more use of the horn and flash functionality; and (4) patient and careful driving \cite{taubman2004multidimensional}. Similarly, Huysduynen categorized driving style as angry driving, anxious driving, dissociative driving, distress-reduction driving and careful driving style \cite{van2015measuring}. Horswill \textit{et al.} provided a valuable distinction between skill and style in the context of driving behaviors \cite{horswill1999}. Hong \textit{et al.} \cite{hong2014smartphone} differentiated styles in terms of defensiveness, as well as by propensity for violation of rules. Scherer defined driving style in terms of comfort \cite{scherer2015driver}. Lee \textit{et al.} \cite{lee2004comprehensive} analyzed lane changes as a function of its severity (degree to which the vehicle in the destination lane was cut off), urgency (how soon the lane change was needed), and type classification for the full population of 8,667 lane changes.

\textit{We focus on driving style based on degree of defensiveness.} 

Driving style is a ``humanized driving'' quality \cite{Here360}. Hence, most of the driving style literature relates to understanding and modeling human driver behavior, in very specific traffic situations or contexts, like lane changing \cite{lee2004comprehensive, salvucci2004inferring, mandalia2005using}, intersection crossing \cite{hong2014smartphone, banovic2016modeling, elhenawy2015modeling}, car following \cite{brackstone1999car}, and in terms of driving actions specific to those contexts (e.g., throttle and braking level, turning) and features thereof (e.g. rate of acceleration, rate of deceleration, maximum speed in a time window).\emph{ We define driving defensiveness in our work as an aggregate of driving features in various driving scenarios}. Therefore, in our study, we present a combination of all of the aforementioned traffic conditions and scenarios to our participants.

Research on driving styles has been extended to autonomous cars in two forms. One body of work includes exploratory studies on understanding how explicitly-defined driving styles relate to comfort \cite{scherer2015driver}. The second body of work encompasses research on ways to teach an autonomous car how to drive from human demonstrations \cite{abbeel2004apprenticeship, ziebart2008navigate, kuderer2015learning,silver2010learning}. Both these groups assume that an autonomous car should learn their own user's driving style or driving behavior. 

\section{Methods}\label{sec:methods}

\subsection{Hypothesis}
Because being a passenger is a different experience than being a driver, we hypothesize that:

\noindent\textbf{H.}
\emph{Users of autonomous cars prefer a driving style that is significantly different than their own.}
 
\subsection{Study Design}
In order to test our hypothesis, we leverage a driving simulator, and let users experience and evaluate autonomous cars with different driving styles, including their own style (without their knowledge). 

We conducted a study in two parts. In the first part we collected driving data of participants in a simulation environment, so that we could let them experience their own style in the second part of the study. 

\subsection{Manipulated Variables}
We manipulated the driving styles of autonomous cars at four levels of defensiveness: \be{aggressive}, \be{defensive}, \be{own style}, and a \be{distractor style} (a different participant's style). Users did not know if any of the styles were their own. Likewise, we also consciously avoided the use of the phrase ``driving style'' anytime during the studies, as well as, in the pre-study screening.

We define the \emph{defensiveness} of the style objectively, as a function of several driving features (e.g., distance to other cars -- the larger the distance, the more defensive the driving). We use features informed by existing literature. We describe them in \sref{sec:features}. 

We created the aggressive and the defensive styles of driving by demonstration, and then validated these styles using our driving features (see our Manipulation Check \sref{sec:manipulationcheck}). 

\subsection{Simulator and Driving Tasks}
We conducted both parts of the study in a simulation environment. Our simulation environment consisted of a standard classroom projection screen and table in front of the screen fitted with Logitech G920 steering wheel, brake, and gas pedal. We used the OpenDS driving simulation software \cite{opends2016} for running each of the driving simulations. The simulation platform was set up on a standard PC augmented with NVIDIA GeForce GTX 1070 and was hidden from the participants' view.  

In the first part, the participants drove on a 9.6 mile long test track that consisted of 14 different driving tasks designed using the City Engine software (\figref{fig:track}). 

We define a \emph{driving task} as a sequence of driving maneuvers in response to specific traffic conditions. For each task there are two to three simulated traffic conditions that resemble everyday traffic, so as to elicit natural driving behavior from the participant. 

In the second part of the study, the participants experienced 6 of these 14 tasks, each performed by autonomous cars of four different styles.

\subsection{Procedure}

Before the driving session in part one of the study, we familiarized participants to the driving simulator. We asked each participant to practice on two different test tracks until they felt that they were driving as they would in their everyday driving. The first track had several traffic signals and turns, and second one was on congested city roads with several traffic cars.  Their driving was assisted by a voice navigation. There were road signs for speed change zone, speed limit, sharp turns, entry to expressway and exit from expressway.  We instructed the participants to drive as they would on actual roads and to treat the speed limits the way they would in their usual driving. This practice session lasted 5-10 minutes for each participant.

Participants then began the first part of the study, which consisted of 15-20 minutes of driving along the 14 tasks-test track, followed by a 10 minute interview.

\begin{figure*}[t!]
    \centering
    \includegraphics[width=\textwidth]{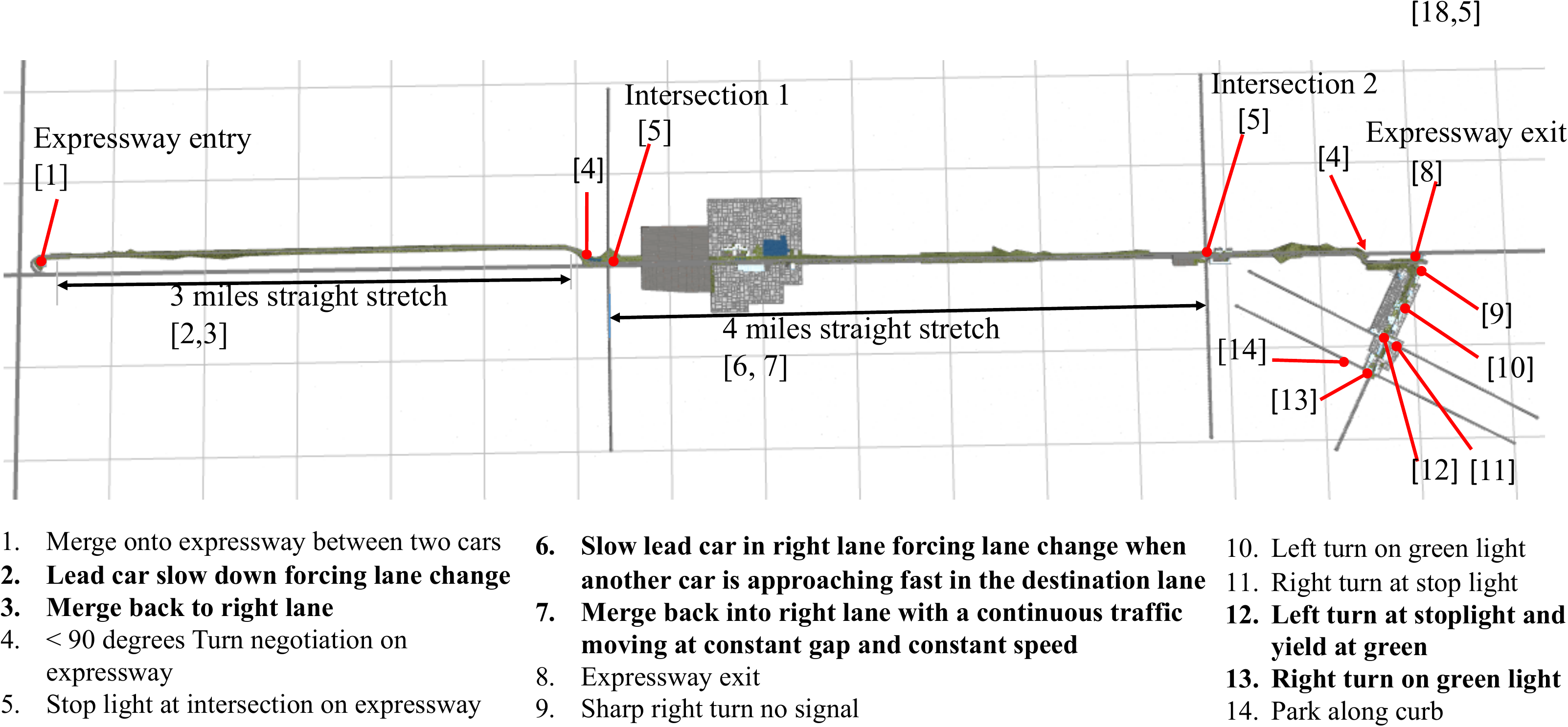}
    \caption{Designed track: Tasks (shown in the list below the figure) are indicated in square brackets. Total road stretch is 9.6 miles.}
    \label{fig:track}
\end{figure*}

In the second part of the study, the autonomous cars performed six tasks (combined into four test tasks) from this list with the participant as a passenger, shown in bold letters on the list in \figref{fig:track}. To simplify, we combined the second and the third tasks in the list, i.e., lead car slows down forcing lane change and merge back to right lane into a single test task, which we refer to as Task 1 in the rest of the paper. Likewise, we combined the sixth and the seventh tasks into a single test task, called Task 2 in the rest of the paper. Thus, each autonomous car performed four test tasks in total.
Two of the test tasks were on the expressway and lasted approximately 4 minutes for each style and the other two tasks on the inner city roads were shorter than 2 minutes.

After the participants had driven in an autonomous car of each driving style for each of the test tasks, we conducted a short interview-based survey with each participant.

\subsection{Dependent Measures}
\label{sec:dependent}
\prg{Perceived similarity to real driving}
In the first part of the study we conducted a post-driving open-ended interview with the participants to understand whether the manual driving in the simulation environment resembled their everyday driving. We asked three questions in this interview, each followed by a request for more elaboration. We asked the following questions in the interview: 
\begin{enumerate}
 \item Did you enjoy the drive?
 \item Are there any positive or negative aspects of the simulation environment, the driving controls and the traffic conditions that you would like to mention?
 \item  On a scale of +3 to -3 \cite{ashrae2010}, please rate how similar or different is this experience from your daily driving?
\end{enumerate}  

\prg{Open-ended responses} In the second part we asked each participant to think aloud about their emotions and feelings as they were experiencing autonomous driving. 

\prg{Main subjective measures: Preference and perceived similarity to own style} After a participant had experienced each autonomous style for a given task, we conducted an interview-based survey. We asked the participants to rate each style of driving for \emph{comfort}, \emph{safety}, \emph{preference for everyday use}, and \emph{similarity with their own driving} on 7 point Likert scale.

\prg{Main objective measures: Driving style features and overall defensiveness}
\label{sec:features}
We measured the user's style quantitatively using task specific driving features, derived from existing literature. We carefully considered the contexts and subject demographics of each of these existing studies to ensure as much similarity in the context as possible with our study. 

For car following, lane changing, and return to preferred lane, we selected the features described by Lee \textit{et al.} in ``A Comprehensive Examination of Naturalistic Lane-Changes'' \cite{lee2004comprehensive}. This study analyzed the largest naturalistic lane change dataset and specifically labelled lane change data resulting from the slowing down of the leading car. The speed range of 45 mph to 55 mph matches our driving conditions. Their dataset consisted 8667 lane changes over 23,949 miles of driving from 16 commuters of age group 20 to 60. They studied car following, lane changing, and return of preferred lane in terms of distance, time to collision, and relative speed classified by severity and urgency of lane change.

The features for tasks like turning at the intersection with a green light or stop light were derived from our preliminary interview with the participants and from Hong \textit{et al.} \cite{hong2014smartphone} and Banovic \textit{et al.} \cite{banovic2016modeling}. 

\begin{table*}
    \centering
    \begin{tabular}{p{2in}|p{4in}}
         Features & Definitions  \\
         \hline
         Mean Distance to Lead car & During car following (with 200 meters distance) the average distance between middle of the driver car and the lead car.\\ 
Mean Time Headway & During car following (with 200 meters distance) average time headway, defined as ratio of Distance headway and speed of the driver car.\\
Time Headway during Lane change & Distance headway divided by the speed of the driver car during lane change.\\
Distance Headway during Lane change & Distance between the middle of the driver car and the lead car during lane change\\
Distance Headway Merge Back & This is the same as Distance Headway during lane change except measured in between driver car and the following car in the destination lane.\\
Braking Distance from the Intersection & The distance from the intersection at which a person starts applying brakes.\\
Time To Stop & Braking distance divided by the speed of the car right before brake is applied.\\
Maximum Turn Speed & Maximum speed of the driver car over a time window during a left turn or a right turn.\\
Speed at the Intersection & Instantaneous speed at the intersection.\\
Average Speed for 20 meters before Intersection & This is the speed of the driver car averaged over a distance range of 20 meters from the intersection.\\
    \hline
    \end{tabular}
    \caption{Features for style classification}
    \label{tab:features}
\end{table*}

Table \ref{tab:features} summarizes all the features for the four driving test tasks. We used \textit{mean distance to lead car}, \textit{mean time headway}, \textit{time headway during lane change}, and \textit{distance headway during lane change} as features for Task 1 and Task 2. Task 1 had an extra feature \textit{distance headway merge back} for scoring the merge back behavior to the right lane.

Task 3 consisted of two sub-tasks (approaching intersection at a stop light and then making a left turn at green ball). We characterized this task with 5 features: \textit{Braking Distance from the intersection}, \textit{Average speed for 20 meters before intersection}, \textit{Time To Stop}, \textit{Speed at the intersection}, and \textit{Maximum turn speed}. 

Task 4 constituted approaching intersection at green ball and then turning right without stopping. The features for this tasks are \textit{Speed at the intersection} and \textit{Maximum Turn Speed}.

We objectively measured a participant's overall driving style in terms of a \be{Defensiveness Score}. We first normalized the feature values across participants for each feature irrespective of the task. We calculated a \textit{Defensiveness Score} for each participant and for each task as the average over all the normalized feature values for that participant and task. We then computed an \textit{Aggregate Defensiveness Score} for each participant by averaging their scores across the four test tasks.

\subsection{Manipulation Check}
\label{sec:manipulationcheck}
We performed a manipulation check on our aggressive and defensive driving styles. We measured the aggregate defensiveness score for each style, plotted on the bottom right of \figref{fig:features}. We found that indeed, the aggressive style was less defensive than the defensive style (lower defensiveness score). We found that 86.67 \% of the users' styles scored higher than the aggressive style, and lower than the defensive style. This suggests that the two reference driving styles created by demonstrations resulted in meaningful representations of aggressive and defensive driving. 

\subsection{Participants}
\prg{Subject Allocation} We opted for a within-subjects allocation because the participants needed to choose a preferred style out of the set of available ones. We randomized the order of the conditions.

\prg{Demographics} We recruited 15 participants consisting of a mix of graduate students and undergraduate students. Before the study we sent out a screening form to each participant in order to ensure a wide distribution of demographics, driving experience and perceived driving behaviors of the participants. We also checked for a valid driving license. 3 of our participants were 30 to 31 years old, the rest of the participants were 18-24 years old.  

The mean driving experience of the participants was 5.46 years with a standard deviation of 4.5 years. Participants had driven an average 214 miles with a standard deviation of 188 miles on the week before they filled out the screening form. 

We asked the participants to give us some information about their perceived driving behavior using the following questions: 1. Please rate if you consider yourself a conservative or an adventurous driver on a 7-point scale, 1 being \textit{conservative} and 7 being \textit{adventurous}. 2. Please rate on a 7-point scale what you like about driving, 1 being \textit{joy of motion} (like feeling the force as you accelerate) and 7 being \textit{comfort of steadiness}. You may like some of both. 3. Rate on a 7-point scale if you think you vary your driving by road conditions, traffic and time availability, 1 being \textit{vary always} and 7 being \textit{I don't vary at all}.  4. Please rate your driving experience from somewhat experienced to very skillful. The purpose of these questions was to acquire some information about the participants' driving styles without explicitly using the term style or in other words give away the original goal of the study. 

Approximately 46 \% of the participants considered themselves well experienced in driving, and 20 \% considered themselves experienced. The rest were equally distributed between somewhat experienced to very skillful. The mean score for perceived conservative-adventurous driving behavior was 3.6. Most of the participants considered themselves to be in the middle of the spectrum. Only one participant considered himself to be conservative.  More participants preferred comfort and steadiness over joy of driving, the average rating being 4.46. The mean rating for variation of driving style in response to environment and traffic was 3, which means most participants believed that they alter their driving behavior according to traffic.\\

\subsection{External Validity and \\Controlling for Confounds}

\prg{Driving environment} We used a simulator and not real autonomous cars. However, we designed a simulation track and traffic conditions so as to elicit natural driving responses.  We also collected participant feedback in the first part of the study on the simulation environment and how their driving behavior in the simulated track related to their actual driving behavior.

\prg{Masking own style}  
One of the major challenges of this work was to ensure that a participant could not recognize his or her driving style from simulation peculiarities like scenes, traffic and controls. We wanted the participants to only recognize their driving style based on their traffic maneuvers and actions. We took several steps to camouflage the driving data of a participant in the second part of the study:
\begin{itemize}
    \item We retained the traffic conditions and route from the first part of the study while changing the surrounding scenes and traffic cars, such that we can replicate the user's driving while removing the bias of familiar environment.
    \item  We let the participant perform approximately 14 driving actions in the first part of the study and picked only some of these tasks for the second part of the study.
    \item During the second part of the study we presented the tasks in an order different from how they occurred in the manual mode. For example: In the first part, the participants first entered the expressway and performed some driving actions on the expressway and then exited the expressway and performed some more driving maneuvers on the city roads. During the second part, we presented one city road task and one expressway task in an alternate order.
    \item We presented the four styles for each of the test driving tasks in a randomized order, which made it more difficult to consistently recognize one style.
    \item We post-processed the users' driving to remove peculiarities, which we explain below.
\end{itemize}

During our pilot studies we found that due to some peculiarities of simulation environment (over-sensitive steering, less sensitive braking) and the resultant jitter in the driving data, some participants were able to recognize their own driving. For example, a participant mentioned: \emph{``This looks like how I was driving. I had to stop at the intersection because I pressed the brake too early. The brake was tight.'' } 
Idiosyncrasies of the simulator led pilot participants to identify their driving behavior in the second part of the study.
In order to eliminate these peculiarities of the simulation environment, we changed the brake stiffness and steering sensitivity and presented participants with smoothed version of their data in the second part of the study. 

 \begin{figure*}[t!]
    \centering
    \includegraphics[scale=0.4]{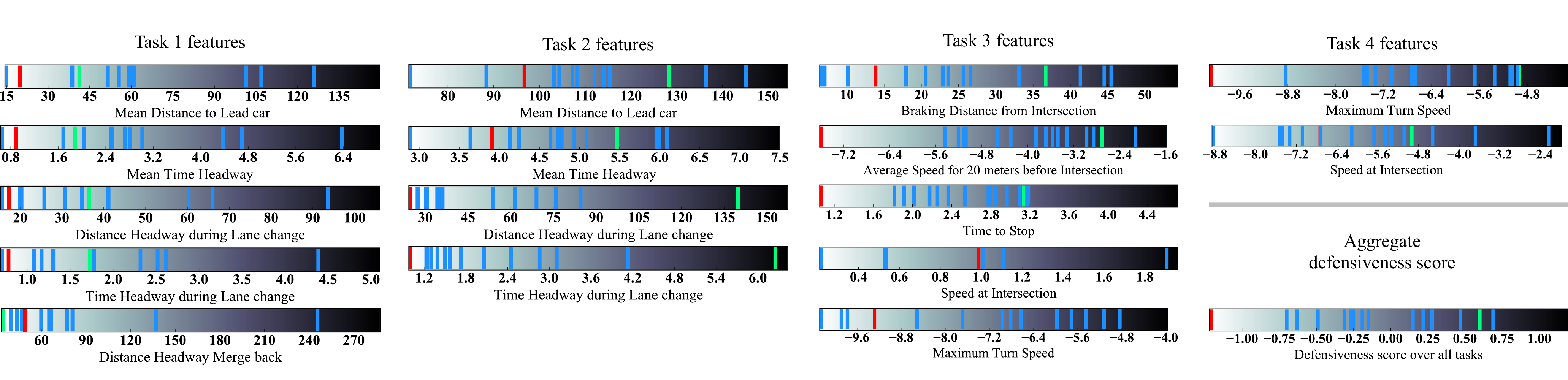}
    \caption{Participants' feature distribution}
    \label{fig:features}
\end{figure*}

\subsection{Trajectory Smoothing}

\begin{figure}[t!]
    \centering
    \includegraphics[scale = 0.8]{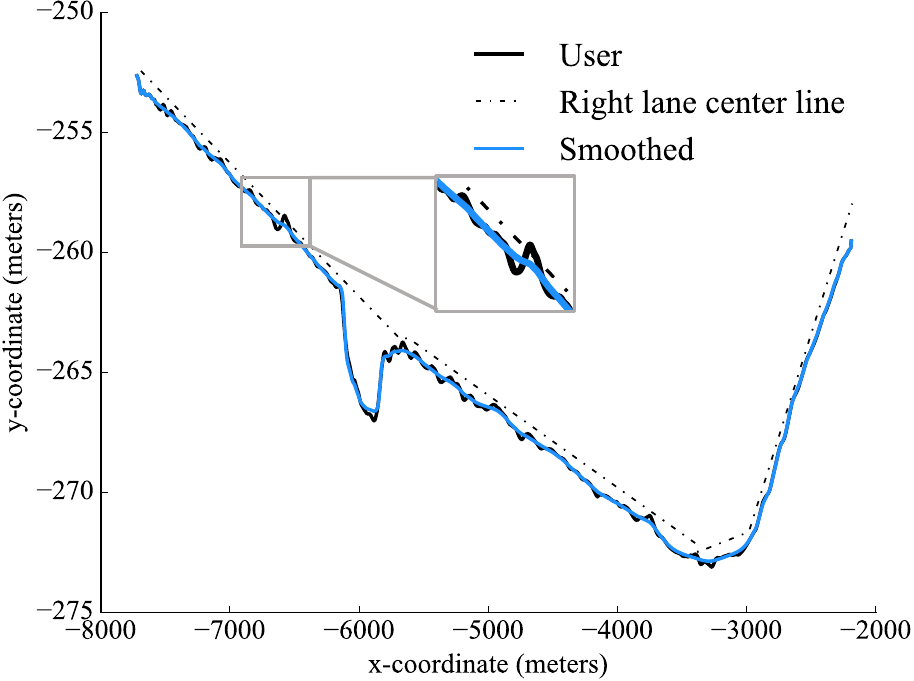}
    \caption{Smoothed trajectory compared to original trajectory of task 1 of one participant at 15 \% smoothing} 
    \label{fig:smoothed_trajectory}
\end{figure}

We filtered the driving trajectories to eliminate idiosyncrasies that make the trajectory instantly recognizable. 

We applied a Bilateral Filter \cite{tomasi1998bilateral} to reduce the lateral variance (or equivalently, the variance of the lateral displacements from the center of the lane) of the trajectories. By affecting only the lateral components of the trajectory, this filtering preserves distance between the cars. We applied filtering only to the stretches of the trajectory on the expressway. 

\figref{fig:smoothed_trajectory} shows a smoothed trajectory for one participant. It has 15 \% lower lateral variance than the original trajectory.



\section{Results} \label{Results}

\subsection{Simulation Realism }

In the first part of the study, in addition to collecting user driving data, we also wanted to ensure that this driving data corresponded to participants' everyday driving as much as possible. We conducted a post-driving interview, as described in the Dependent Measures subsection (\sref{sec:dependent}). Here we present the results of the interview. 

The rating mode for similarity between driving on road and driving in our study simulator was +1 on -3 to +3 scale.  Four participants gave a rating of +2. Some of their positive comments were: ``Not considering the room environment and just looking at the simulation graphics and the car it was pretty much the same environment as real. I would give +3 for surrounding traffic conditions". Other participants said that they felt relaxed in the simulator environment and that they could drive cautiously as they would in real traffic. 

One participant who rated the driving experience similarity -2 complained about the lack of motion feedback in the system. This is the same participant who gave high rating for \textit{joy of motion} in the screening question. However, no other participant had the same concern and got well-adjusted to the simulation environment. 

Most of the participants who rated +1 to -1 found steering re-centering or brake insensitivity difficult. We also received quite opposite feedback from two participants when they compared their everyday driving to the simulator driving. For example, one participant mentioned ``It felt real. It was something I could get used to after driving a while. The gas and brakes were more sensitive than my car''. Another participant felt that the brakes were excellent, different from regular car. 

One participant reported that she was so immersed after driving for a while, that she caught herself turning her head back to check for oncoming traffic in the destination lane. We found that participants with one or less years of driving experience could not use the simulation environment properly. Overall, the ratings and the comments supported that the simulator conditions are not \emph{too} far from real conditions.

\subsection{Feature Distribution for Participant Styles}
We define driving style in terms of features mentioned in the \sref{sec:methods}. 
 
\figref{fig:features} shows, for each task, feature, and participant, what the participant's feature value was for that task (blue marks). The figure also shows the aggressive style values in red and defensive style values in green.

Higher negative values correspond to more aggressive behavior. All the feature values are arranged from aggressive on the left to defensive on the right. However, for features like speed where lower values mean more defensive we show and use the negation of these features. 

The bottommost plot to the right shows the aggregate defensiveness score. This score is derived from the normalized feature values.
60 \% of the participants are within 0.75 standard deviation aggressive and 40 \% within 0.75 standard deviation defensive.  Only two of the participants were more defensive than the autonomous defensive car, one of them being very close to the defensive car in the score. 
\begin{quote}
\emph{When looking at the aggregate defensiveness, most participants lie between the aggressive and defensive styles. }
\end{quote}

There are, however, exceptions, but for particular features in particular tasks.
For task 1, several participants were more defensive than the defensive autonomous car. For the last feature of task 1, Distance Headway Merge Back, the aggressive car was not as aggressive as several participants and even our defensive car. In task 2, the aggressive and the defensive autonomous cars enclosed a middle section of the spectrum for Mean Distance Headway and Mean Time Headway. In other words, several participants were more aggressive and more defensive than the aggressive and defensive autonomous cars respectively. This is because these features were measured during car following over a long time span and are expected to have wider distributions than features characterizing instantaneous actions. 

\begin{figure}[t!]
    \centering
     \includegraphics[scale = 0.5]{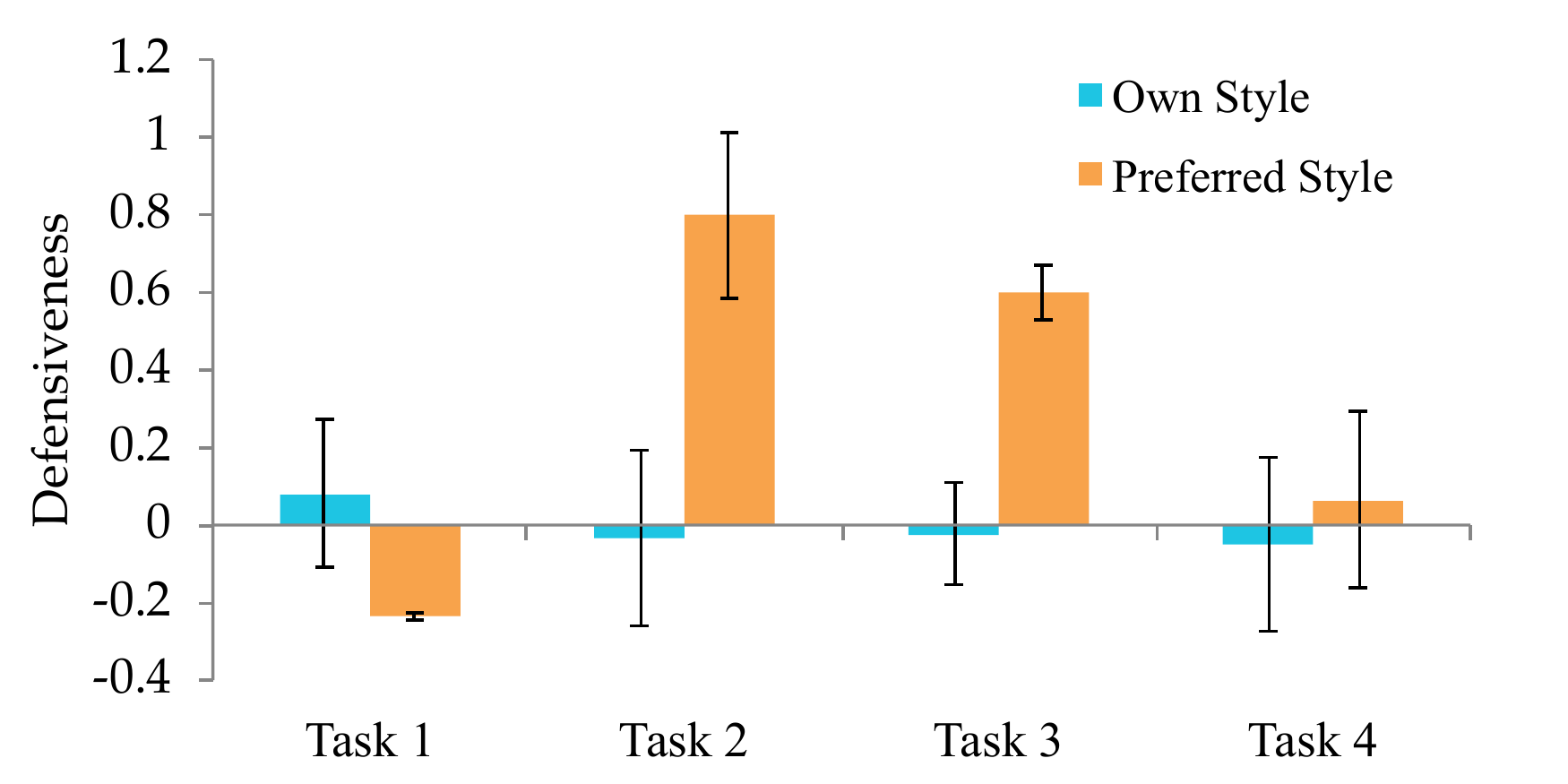}
     \caption{Mean Defensiveness Score Across Participants. The corresponding scores of aggressive and defensive autonomous cars are Task 1: (-0.768, -0.222) Task 2: (-0.885, 1.325), Task 3: (-1.82, 0.766) and Task4: (-1.49,0.72).} 
     \label{fig:user preference}
 \end{figure}

\subsection{Preferred Style in Relation to Own Style}
\label{sec:preferredvsactual}

We asked participants to rate how much they would prefer driving with each style, for each task. We refer to the highest rated style(s) as the participant's preferred style(s).

Our main finding is that overall, users preferred a different style than their own. A total of 9 out of 15 participants preferred a different style than their own on at least one of the tasks. A matched pairs $t$-test comparing actual and preferred defensiveness score showed a significant difference ($t(1,60)=-2.58$, $p=.0121$), supporting our hypothesis. Here, whenever a user's highest rating was for multiple styles as opposed to a single one, we included each preferred style as a data point. 

\begin{quote}
\emph{Overall, people prefer a significantly more defensive style than their own.}
\end{quote}

We also investigated how this breaks down by task, and only found significant effects on the $2^{nd}$ and $3^{rd}$ tasks. See \figref{fig:user preference} for comparison between average preferred style and own style of our participants for each of the four tasks. For task 1 we note that several participants were more defensive than other autonomous styles presented to them. However, they still preferred our defensive style, which explains that the average choice was more aggressive than the participants' own style. 

Interestingly, some participants did not perceive the extra defensive nature of their own style in task 1 positively. One participant mentioned about their own style that ``In this one I felt like we gave a lot of room, more than I would have probably.'' (ironically, since they did \emph{exactly} that). Two other participants made similar comments about their own lane changing behavior. Besides, a few participants also considered driving features beyond the ones we accounted for. 

For task 2 and task 3 the defensive autonomous car was more aggressive than only none to three participants across all features and it was more defensive than the rest of the population by a major margin, in features like Distance Headway and Time Headway During Lane Change. 

The task had a significant effect on the difference ($F(3,58)=4.13$, $p=.0101$), suggesting that people's preferences for a driving style are not consistent, but rather \emph{change based on the context.} This motivates future research on predicting the desired driving style not just based on the individual, but also based on the current driving context.

 \begin{figure}[t!]
    \centering
     \includegraphics[width=\columnwidth]{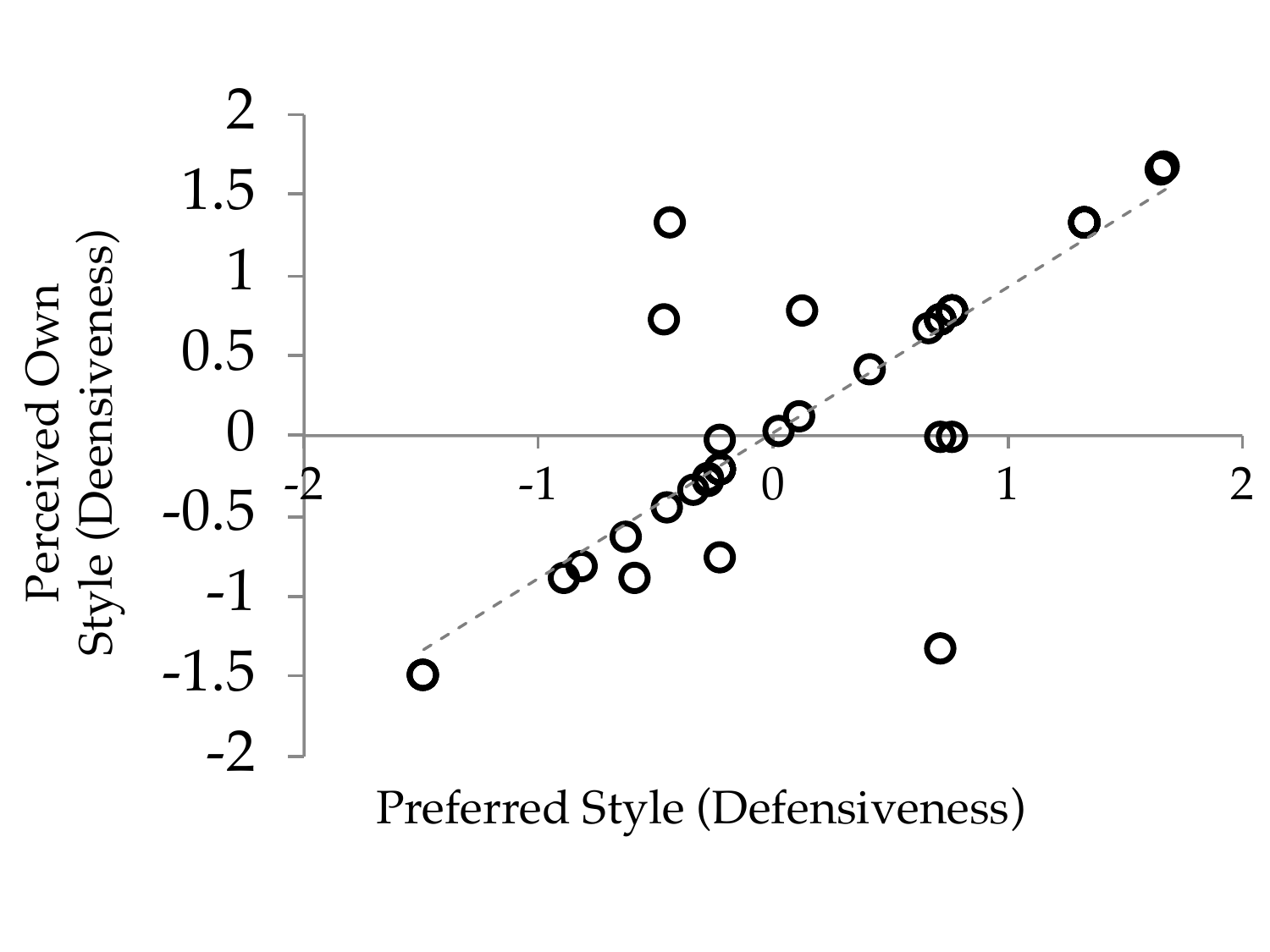}
     \caption{Scatter plot showing correlation between the style that users \emph{thought} was their own and the style that they chose as their preferred.}
     \label{fig:perceived_preferred}
 \end{figure}
 \begin{figure}[t!]
    \centering
     \includegraphics[width=\columnwidth]{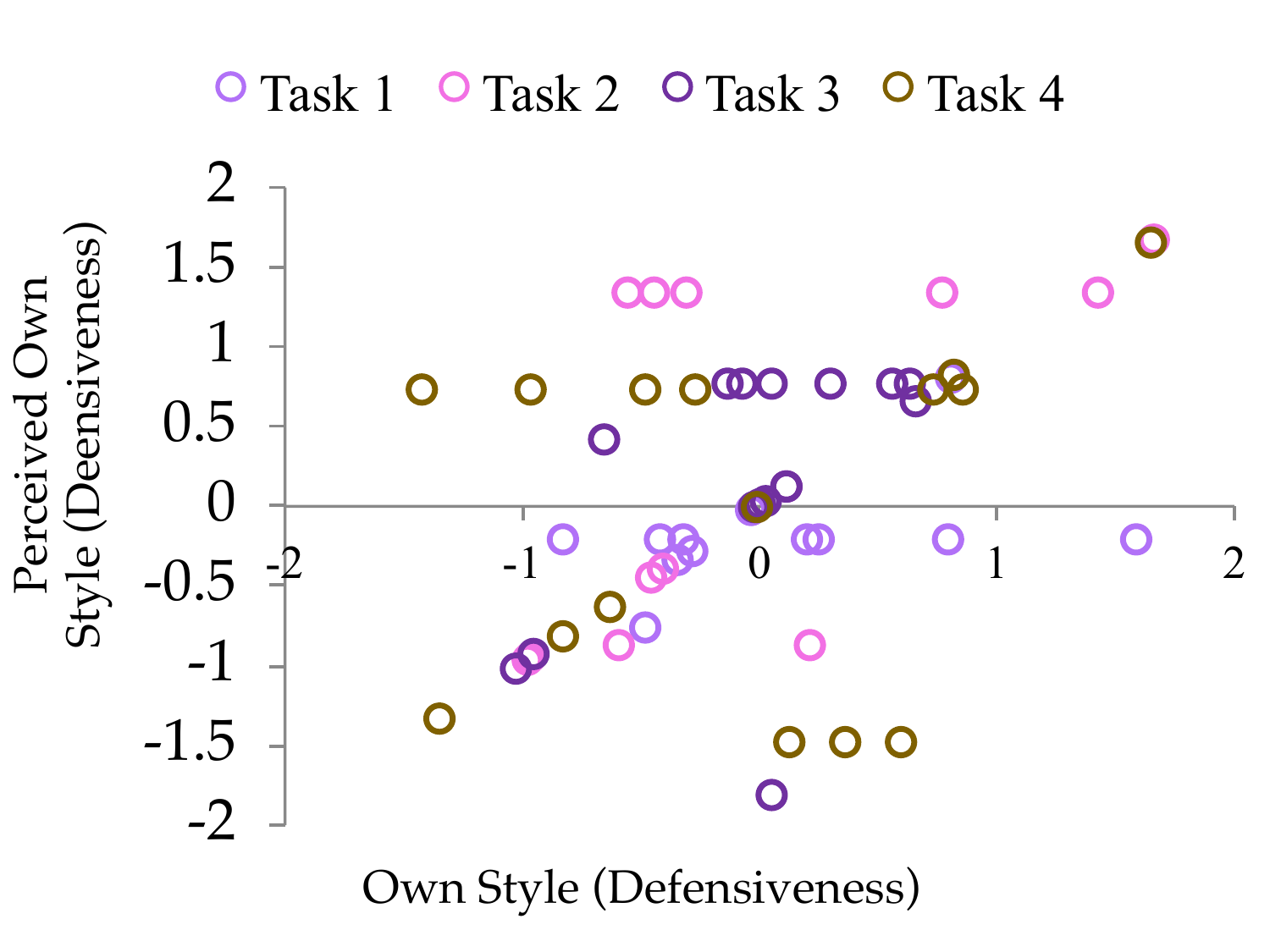}
     \caption{Scatter plot showing little correlation between own style and perceived own style: users did not tend to identify their own style correctly. as evidenced by the off-diagonal points.}
     \label{fig:perceived_own}
 \end{figure}

\subsection{Perceived Own Style \\in Relation to Actual Own Style}

We also asked participants to rate each style in terms of similarity to their own. From this, we learned what participants \emph{perceived} their own style to be. 

We found that even though participants did not pick their \emph{actual} style as their preferred (Sec. \ref{sec:preferredvsactual}), participants did tend to prefer their \emph{perceived} style. On each task, between 80 and 93\% of participants opted for the same style as the one they \emph{thought} was the closest to their own (and sometimes rated other styles as well as equally good). We found a significant correlation between the perceived own and preferred styles, $r(58)=.86$, $p<.0001$.  \figref{fig:perceived_preferred} shows a scatter plot of preferred style by perceived style, with many points on the diagonal representing users who preferred driving in the style they thought was (closest to) their own.

However, even though the majority participants thought that they were picking their own style, they really were not. A total of 46 to 67\% participants on each task did not correctly identify their actual own style, and the correlation between perceived and actual defensiveness score was only $r(56)=.40$ across tasks. \figref{fig:perceived_own} paints a different picture from \figref{fig:perceived_preferred}: it plots the perceived style against the \emph{actual} own style, showing many off-diagonal points, representing users who did not correctly identify their style.

In task 1 we see that several participants perceived themselves to be slightly more aggressive irrespective of their actual style. Likewise, both for task 2 and task 3 several participants perceived themselves to be more defensive irrespective of their actual style.

\begin{quote}
\emph{Participants tended to prefer the style that they thought was their own, but in fact that style had little correlation to their actual own style. }
\end{quote}

\section{Discussion}

\noindent\textbf{Summary.} We hypothesized that users of future autonomous cars would prefer a driving style that is significantly different than their own. We conducted a user study in a driving simulator to test our hypothesis. We found that users preferred a more defensive style than their own. This echoes the finding from prior work \cite{horswill1999} that when people are not in control of the driving they prefer lower speeds -- autonomous cars are one instantiation of not being in control of the driving.

Interestingly, over 80\% of users preferred the style that they \emph{thought} was their own, but many times they were incorrect in identifying their own style. These results open the door for learning what the user's  preferred  style  will  be,  but caution against getting driving demonstration from the user, since people can drive like they do, not like they want to be driven.

\noindent\textbf{Limitations and Future Work} 
Our work is limited in the following ways:
\begin{itemize}
\item \textbf{Limited driving style features}. Following the most common conventions, we have only characterized style in terms of defensiveness. We also inherited from previous studies the feature choices in defining driving styles. 

\item \textbf{Limited driving style choices}. We presented participants with limited options along the spectrum of defensiveness and found that they preferred a style more defensive than their own. However, we did not learn the style they actually desired, only the best out of our few options.

\item \textbf{Limited fidelity of simulation environment}. Our simulation environment does not provide motion feedback, which may limit the users' perception of speed. Although the interview results validated that participants' perception of the driving styles are sufficient, experiment results in a higher fidelity simulation environment might be more accurate. 

\end{itemize}

Given the encouraging results from our findings presented here, we believe that it is worthwhile to test more diverse feature choices and  driving style representations in a higher fidelity setting. It is also worthwhile exploring what features users' consider when they evaluate autonomous driving styles. These experiments will provide a comprehensive evaluation of the study presented in this paper.

Going further, we are excited to investigate how we might learn a deviation from the user's driving style that is predictive of how they actually want to be driven, and explore new learning techniques that can augment user demonstrations with other types of user input and guidance. 

\section{Acknowledgements}
We would like to thank our post-doctoral colleague Santosh Chandrasekhar and undergraduate researcher Joseph Stansil for their contribution in system set-up and pilot studies for this research. This work is partly supported by the Berkeley Deep Drive Center, the Center for Human-Compatible AI, and CITRIS.

\bibliographystyle{abbrv}
\bibliography{driverpreference}  

\end{document}